\pdfoutput=1

\documentclass[11pt]{article}

\usepackage{EMNLP2023}

\usepackage{times}
\usepackage{latexsym}

\usepackage[T1]{fontenc}

\usepackage[utf8]{inputenc}

\usepackage{microtype}

\usepackage{inconsolata}

\usepackage{comment}
\usepackage{graphicx}
\usepackage{subfigure}
\usepackage{xcolor}
\usepackage{enumitem}
\usepackage{amsmath}
\usepackage{amssymb}
\usepackage{xcolor}
\usepackage{multirow}

\newcommand{\ie}{\textit{i.e.}}
\newcommand{\eg}{\textit{e.g.}}
\newcommand{\erm}{\textsc{erm}}
\newcommand{\kd}{\textsc{kd}}
\newcommand{\ls}{\textsc{ls}}
\newcommand{\mrpc}{\textsc{mrpc}}
\newcommand{\qnli}{\textsc{qnli}}
\newcommand{\sst}{\textsc{sst-2}}
\newcommand{\mnli}{\textsc{mnli}}
\newcommand{\glue}{\textsc{glue}}
\newcommand{\distilbert}{\textsc{d}istil\textsc{bert}}
\newcommand{\bert}{\textsc{bert}}
\newcommand{\nlp}{\textsc{nlp}}
\newcommand{\colorgreen}{\textcolor[rgb]{0,.6,0}}
\newcommand{\colorred}{\textcolor[rgb]{.75,0,0}}

\title{Knowledge Distillation $\approx$ Label Smoothing: Fact or Fallacy?}

\author{Md Arafat Sultan\\ IBM Research AI \\ arafat.sultan@ibm.com}

\begin{document}
\maketitle
\begin{abstract}
Originally proposed as a method for knowledge transfer from one model to another, some recent studies have suggested that knowledge distillation (\kd{}) is in fact a form of regularization.
Perhaps the strongest argument of all for this new perspective comes from its apparent similarities with label smoothing (\ls{}).
Here we re-examine this stated equivalence between the two methods by comparing the 
predictive confidences of the models they train.
Experiments on four text classification tasks involving models of different sizes show that: (a)~In most settings, \kd{} and \ls{} drive model 
confidence in completely opposite directions, and (b) In \kd{}, the student inherits not only its knowledge but also its 
confidence from the teacher, reinforcing the classical knowledge transfer view. 
\end{abstract}

\section{Introduction}
\label{section:introduction}
Knowledge distillation (\kd{}) was originally proposed as a mechanism for a small and lightweight \textit{student} model to learn to perform a task, \eg{}, classification, from a higher-capacity \textit{teacher} model \cite{hinton2014distilling}.
In recent years, however, this view of \kd{} as a knowledge transfer process has come into question, with a number of studies finding that it has features characteristic of a regularizer \cite{yuan2020revisiting,zhang2020self,dong2019distillation}.
Perhaps the most direct argument for this new position comes from \citet{yuan2020revisiting}, who show that \kd{} has interesting similarities with label smoothing (\ls{}) \cite{szegedy2016rethinking}. 
One of their main conclusions, that ``\kd{} is a type of learned \ls{} regularization,'' has subsequently been reiterated by many \cite{menon2021statistical,chen2021crosslayer,shen2021label,chen2021robust} and remains an important part of our understanding of \kd{} to this day.

This is a rather curious suggestion, however, given the starkly different purposes for which the two methods were originally developed: \ls{} was designed \textit{``for encouraging the model to be less confident''} \cite{szegedy2016rethinking} by imposing a uniform prior on the target distribution;
the resulting new distribution is intended to simply prevent overfitting without revealing any new details about the task.
\kd{}, on the other hand, was designed to explicitly teach the student better inter-class relationships for classification, so that 
\textit{``a BMW, for example, may only have a very small chance of being mistaken for a garbage truck, but that mistake is still many times more probable than mistaking it for a carrot''}
\cite{hinton2014distilling}.

Such different motivations of the two methods prompt us to re-examine the claim of their equivalence in this paper, where we find the underlying argument to be essentially flawed.
With a growing dependence in key areas like \nlp{} on increasingly large models, \kd{} and other model compression methods are becoming more relevant than ever before \cite{wang2023self,taori-etal-2023-alpaca,dettmers-etal-2023-qlora}.
It is therefore crucial that we also interpret such methods correctly to help steer their future research in the right direction, which is a primary goal of this work.

\kd{} and \ls{} do indeed share the common property that they both substitute one-hot training labels with \textit{soft} target distributions.
The key question for us, however, is whether the soft teacher estimates in \kd{} simply regularize training like the smoothed labels in \ls{} do, or if they present a detailed and finer-grained view of the task, as originally intended.
\citet{yuan2020revisiting} argue in favor of the former based on observations such as (a) similarities in the cost functions of \kd{} and \ls{}, and (b) strong generalization of distilled image classifiers independent of the level of expertise of their teachers.

Here we first analyze their arguments and point out why they may be inadequate for a ``\kd{} $\approx$ \ls{}'' assertion (\S{\ref{section:ls-view-of-kd}, \ref{section:deeper-look}}).
Then we present empirical evidence on four text classification tasks from \glue{} \cite{wang2019glue} that \textbf{\textit{\kd{} lacks the defining property of \ls{}}} in most settings.
Concretely, relative to standard empirical risk minimization (\erm{}) over one-hot expert labels, \ls{} \textit{by design}
increases the uncertainty (entropy) in the trained model's posterior over the associated categories.
In other words, it always reduces the model's confidence in its predictions, as expected.
The effect of \kd{} on the student's posterior, on the other hand, is a function of the relative capacities of the teacher and the student:
In the more common setting where a larger teacher is distilled into a smaller student and also in self-distillation, \kd{} almost always reduces entropy over \erm{}, while also improving accuracy.
It is only when the student is larger than its teacher
that we see a small increase in entropy---still much less than in \ls{}---along with a smaller gain in accuracy.
These results clearly show that unlike \ls{}, \kd{} does not operate by merely making the trained model less confident in its predictions.

The following is a summary of our contributions:
\begin{itemize}[topsep=2pt,itemsep=-1pt]
    \item We study the inner workings of \kd{} in the relatively under-explored modality of language, specifically on text classification tasks.
    \item We show that \kd{} lacks the defining property of \ls{} regularization, rebutting the claim that the former is a form of the latter.
    \item We demonstrate how the specific selection of the teacher model determines student confidence in \kd{}, providing strong support for its classical knowledge transfer view.
\end{itemize}

\section{The Argument for \kd{} $\approx$ \ls{}}
\label{section:ls-view-of-kd}
Let us first take a look at the main arguments, primarily from \citet{yuan2020revisiting}, for why \kd{} should be considered a form of \ls{}.
Given a training instance $x$, let $q(k{\mid}x)$ be the corresponding one-hot ground truth distribution over the different classes $k \in \{1, ..., K\}$.
\ls{} replaces $q(k{\mid}x)$ with the following smoothed mixture as the target for training:
\begin{equation*}
\label{equation:ls-target}
    q_{\scriptscriptstyle ls}(k{\mid}x) = (1 - \alpha)q(k{\mid}x) + {\alpha}u(k)
\end{equation*}
where $u(k)=1/K$ is the uniform distribution over the $K$ classes.
\citet{yuan2020revisiting} show that optimizing for $q_{\scriptscriptstyle ls}$ is equivalent to minimizing the following cost function:
\begin{align}
\label{equation:ls-cost-function}
    \mathcal{L}_{LS} = (1 - \alpha)H(q,p) + {\alpha}D_{KL}(u,p)
\end{align}
where $p(k{\mid}x)$ is the model output for $x$, and $H$ and $D_{KL}$ are the cross-entropy and the Kullback-Leibler divergence between the two argument distributions, respectively.
Note that the $H(q,p)$ term is the same cross-entopy loss $\mathcal{L}_{ERM}$ that is typically minimized in standard empirical risk minimization (\erm{}) training of classifiers.

\kd{}, on the other hand, is defined directly as minimizing the following cost function:
\begin{equation}
\label{equation:kd-cost-function}
    \mathcal{L}_{KD} = (1 - \alpha)H(q,p) + {\alpha}D_{KL}(p^t,p)
\end{equation}
where $p^t(k{\mid}x)$ is the posterior computed by a pre-trained teacher $t$ given $x$.
The similarity in the forms of Eqs. \ref{equation:ls-cost-function} and \ref{equation:kd-cost-function} is one of the main reasons why \citet{yuan2020revisiting} interpret \kd{} ``as a special case of \ls{} regularization.''

A second argument stems from their observation on a number of image classification datasets that a ``teacher-free \kd{}'' method, which employs a smoothed target distribution as a virtual teacher, has similar performance as standard \kd{}.
Finally, some of their empirical observations, where \kd{} improves student accuracy even when the teacher is a weak or poorly trained model, provide support for the more general view of \kd{} as a form of regularization.
Investigation of this general question can be found in others' work as well, \eg{}, in the context of multi-generational self-\kd{} \cite{zhang2020self,mobahi2020self} and early stopping \cite{dong2019distillation}.


Finally, a related line of work view \kd{} as ``instance-specific label smoothing'', where the amount of smoothing is not uniform across categories, but is a function of the input approximated using a pre-trained network \cite{zhang2020self,lee2022adaptive}.
We note that this is a fundamentally different assertion than \kd{} being a form of \ls{} with a uniform prior, since such methods do in fact bring new task-specific knowledge into the training process, with the pre-trained network essentially serving as a teacher model.

\section{A Closer Look at the Arguments}
\label{section:deeper-look}
As already stated, the focus of this paper is on the specific claim by \citet{yuan2020revisiting} that \kd{} is a form of \ls{} with a uniform prior \cite{szegedy2016rethinking}, a primary argument for which is that their cost functions in Eqs.~\ref{equation:ls-cost-function} and \ref{equation:kd-cost-function} have a similar form.
A careful examination of the two functions, however, reveals a key difference.

To elaborate, let $H(.)$ represent the entropy of a probability distribution; it then follows  that \mbox{$\forall{p\in\Delta^n}\backslash\{u\}\mkern3muH(u)\mkern0mu>H(p)$}, where $\Delta^n$ is a probability simplex, $u$ is the uniform distribution and $p$ a different distribution on $\Delta^n$.
Accordingly, minimizing $D_{KL}(u,p)$ in Eq.~\ref{equation:ls-cost-function} unconditionally increases $H(p)$, training higher-entropy models than \erm{}, a fact that we also confirm empirically in \S{\ref{section:experiments}}.

The same is not true for $D_{KL}(p^t,p)$ in Eq.~\ref{equation:kd-cost-function}, however, since $p^t$ is dependent on the specific selection of the teacher model $t$, and $H(p^t)$ can take on any value between 0 and $H(u)$.
Crucially, this means that $H(p^t)$ can be both higher or lower than $H(p)$ if the two respective models were trained separately to minimize $\mathcal{L}_{ERM}$.
Unlike in \ls{}, it therefore cannot be guaranteed in \kd{} that minimizing $D_{KL}(p^t,p)$ would 
increase $H(p)$.
In \S{\ref{section:experiments}}, we show how the selection of the teacher model affects the predictive uncertainty of the student, for different combinations of their sizes.
Based on the same results, we also comment on the teacher-free \kd{} framework of \citet{yuan2020revisiting}.

\section{Experimental Results}
\label{section:experiments}
\noindent 
\textbf{\textit{\color{darkblue} Setup.}}
We use four text classification tasks from \glue{} \cite{wang2019glue} in our experiments:
\begin{itemize}[topsep=1pt,itemsep=-4pt,leftmargin=12pt]
    \item \textbf{\mrpc{}} \cite{dolan2005automatically} asks for a \textit{yes}/\textit{no} decision on if two given sentences are paraphrases of each other.
    \item \textbf{\qnli{}} provides a question and a sentence from \textsc{sq}u\textsc{ad} \cite{rajpurkar2016squad} and asks if the latter answers the former (\textit{yes}/\textit{no}).
    \item \textbf{\sst{}} \cite{socher2013recursive} is a sentiment detection task involving \textit{positive}/\textit{negative} classification of movie reviews.
    \item \textbf{\mnli{}} \cite{williams2018broad} is a 3-way (\textit{entailment}/\textit{contradiction}/\textit{neutral}) classification task asking for whether a premise sentence entails a hypothesis sentence.
\end{itemize}
For each task, we use a train-validation-test split for training, model selection, and evaluation, respectively. 
See Appendix \ref{subsection:appendix-datasets} for more details.

To examine the effect of \kd{} in depth, like \citet{yuan2020revisiting}, we run experiments under three different conditions:
\textbf{\textit{Standard \kd{}}} distills from a fine-tuned \bert{}-base \cite{devlin2019bert} (\bert{} henceforth) classifier into a \distilbert{}-base \cite{sanh2019distilbert} (\distilbert{} henceforth) student;
\textbf{\textit{Self-\kd{}}} distills from a \distilbert{} teacher into a \distilbert{} student;
\textbf{\textit{Reverse \kd{}}} distills from a \distilbert{} teacher into a \bert{} student.
In all cases, the teacher is trained using \erm{} with a cross-entropy loss.

\begin{table}[h]
    \small
    \centering
    \begin{tabular}{cc}
        \textbf{Hyperparameter} & \textbf{Values} \\
        \hline
        Learning rate & $\{1, 2, ..., 7\} \times 10^{-5}$ \\
        $\#$ of epochs & $\{1, 2, ..., 10\}$ \\
        $\alpha$ (Eqs. \ref{equation:ls-cost-function} and \ref{equation:kd-cost-function}) & $\{.1, .2, ..., .9\}$
    \end{tabular}
    \caption{Hyperparameter grid for model selection.}
    \label{table:hp-grid}
\end{table}

For every model evaluated under each condition, we perform an exhaustive search over the hyperparameter grid of Table~\ref{table:hp-grid} (when applicable).
Batch size is kept fixed at 32.
We train with mixed precision on a single NVIDIA A100 GPU.
An AdamW optimizer with no warm-up is used.
In Appendix~\ref{subsection:appendix-optimal-hp}, we provide the hyperparameter configuration that is optimal on validation and is subsequently evaluated on test for each model and task.
We use the Transformers library of \citet{wolf2020transformers} for implementation.

The experiments reported in this section use a \kd{} temperature $\tau=1$, effectively disallowing any artificial smoothening of the target distribution.
In Appendix~\ref{subsection:appendix-high-temp-kd}, we also show that treating $\tau$ as a hyperparameter during model selection does not qualitatively change these results.

\begin{table}[h]
    \vspace{2mm}
    \small
    \centering
    \begin{tabular}{c|ccccc}
         \textbf{Method} & \textbf{\mrpc{}} & \textbf{\qnli{}} & \textbf{\sst{}} & \textbf{\mnli{}} \\
         \hline
         \erm{} & .1123 & .1419 & .1101 & .2698 \\
         \hline
         \ls{} & \colorgreen{.6896} & \colorgreen{.6752} & \colorgreen{.4447} & \colorgreen{.8103} \\
         Standard \kd{} & \colorred{.0548} & \colorred{.1210} & \colorred{.0594} & \colorred{.1887} \\
         Self-\kd{} & \colorred{.0738} & \colorred{.1341} & \colorred{.0977} & \colorgreen{.2758} \\         
    \end{tabular}
    \caption{Test set posterior entropies of different \distilbert{} classifiers. \ls{} increases entropy over \erm{} as expected, while \kd{} almost always reduces it.}
    \label{table:dbstud-entropies-temp1}
\end{table}

\begin{figure*}[t]
\centering
\subfigure[\mrpc{}]{\label{figure:mrpc-dbstud-t1}\includegraphics[width=39mm]{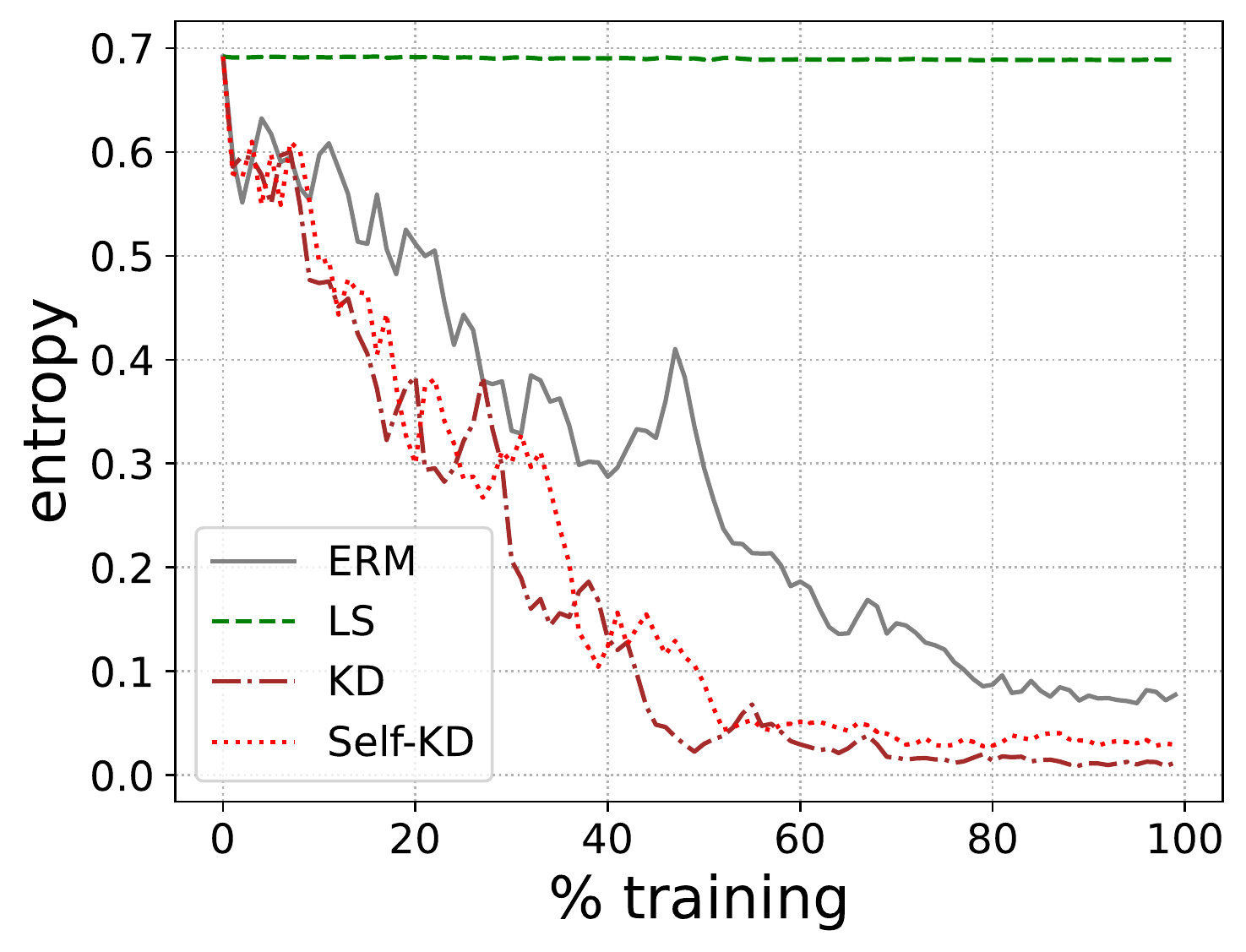}}
\subfigure[\qnli{}]{\label{figure:qnli-dbstud-t1}\includegraphics[width=39mm]{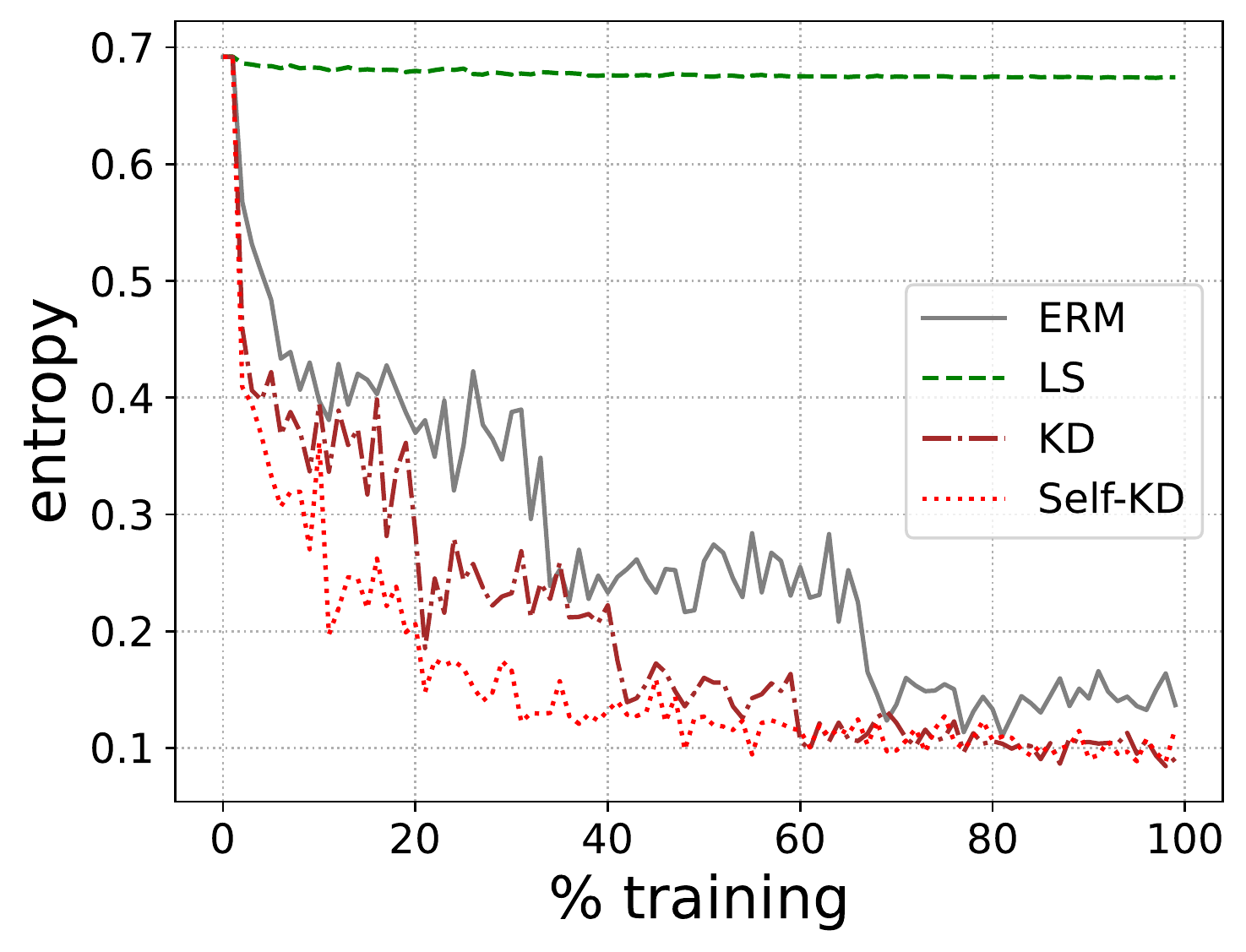}}
\subfigure[\sst{}]{\label{figure:sst2-dbstud-t1}\includegraphics[width=39mm]{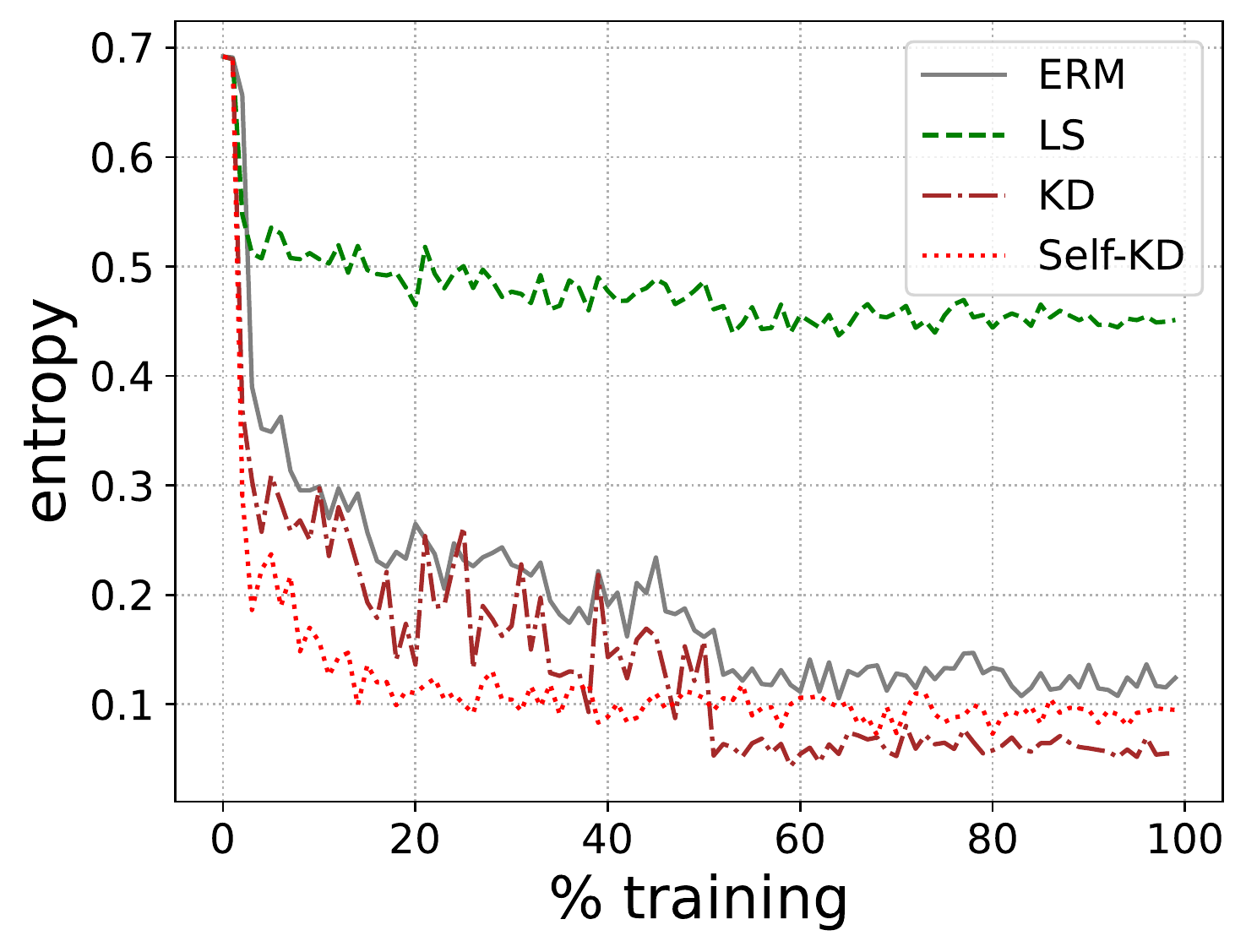}}
\subfigure[\mnli{}]{\label{figure:mnli-dbstud-t1}\includegraphics[width=39mm]{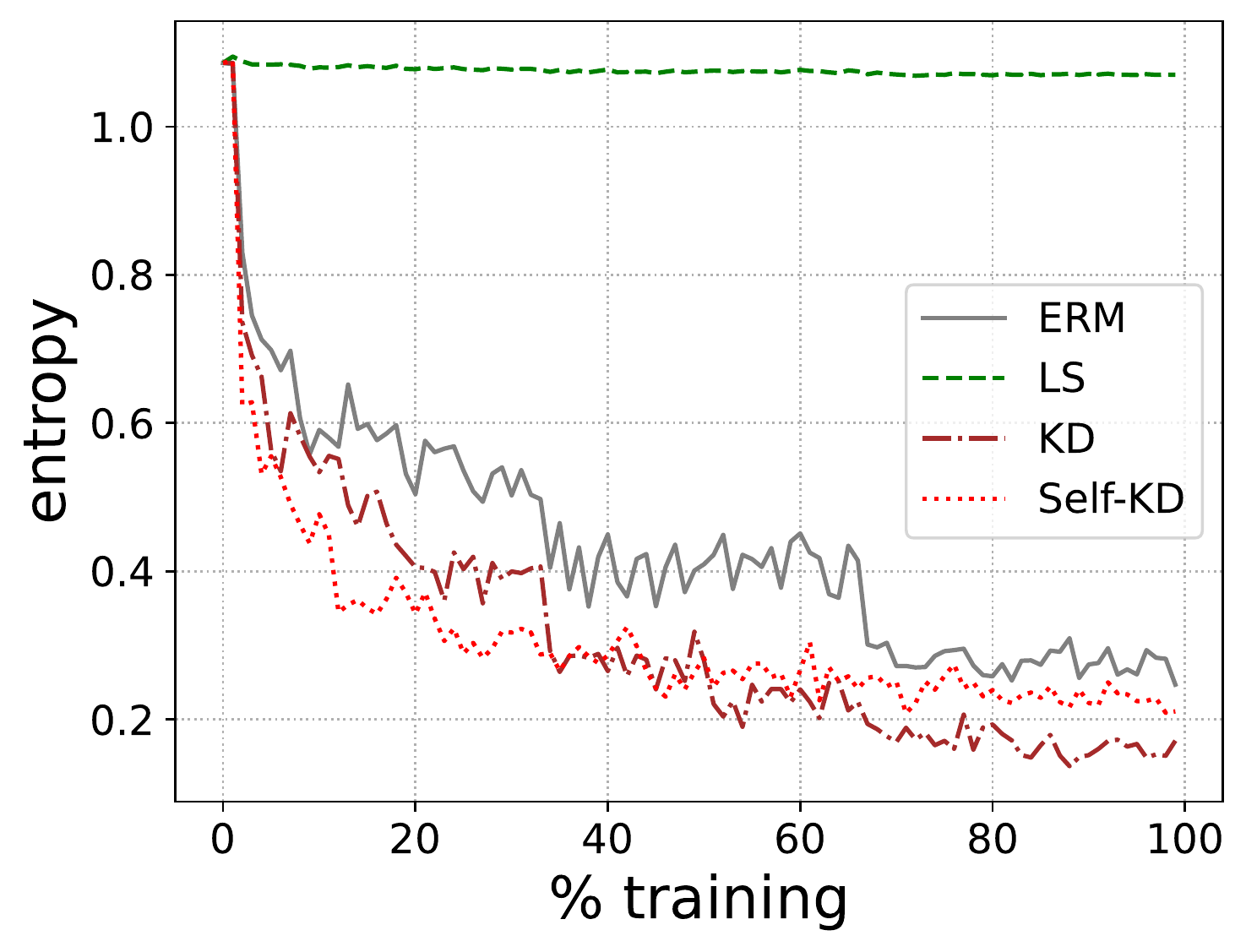}}
\caption{Change in posterior entropy of different \distilbert{} classifiers as training progresses.}
\label{figure:dbstud-entropies-temp1}
\end{figure*}

\noindent
\textbf{\textit{\color{darkblue} Results.}}
Table~\ref{table:dbstud-entropies-temp1} compares the average entropy of posteriors over classes predicted by \ls{} models with those by standard and self-\kd{} students, measured on the four test sets.
As expected, \ls{} increases model uncertainty over \erm{}, in fact quite significantly ($\sim$3--6$\times$).
Both \kd{} variants, on the other hand, bring entropy down; the only exception to this is self-\kd{} on \mnli{}, where we see a very slight increase.
These results have direct implications for the central research question of this paper, as they reveal that the typical effect of \kd{} is a reduction in predictive uncertainty over \erm{}, which is the exact opposite of what \ls{} does \textit{by design}.


\begin{figure*}[ht]
\centering
\subfigure[\mrpc{}]{\label{figure:mrpc-bstud-t1}\includegraphics[width=39mm]{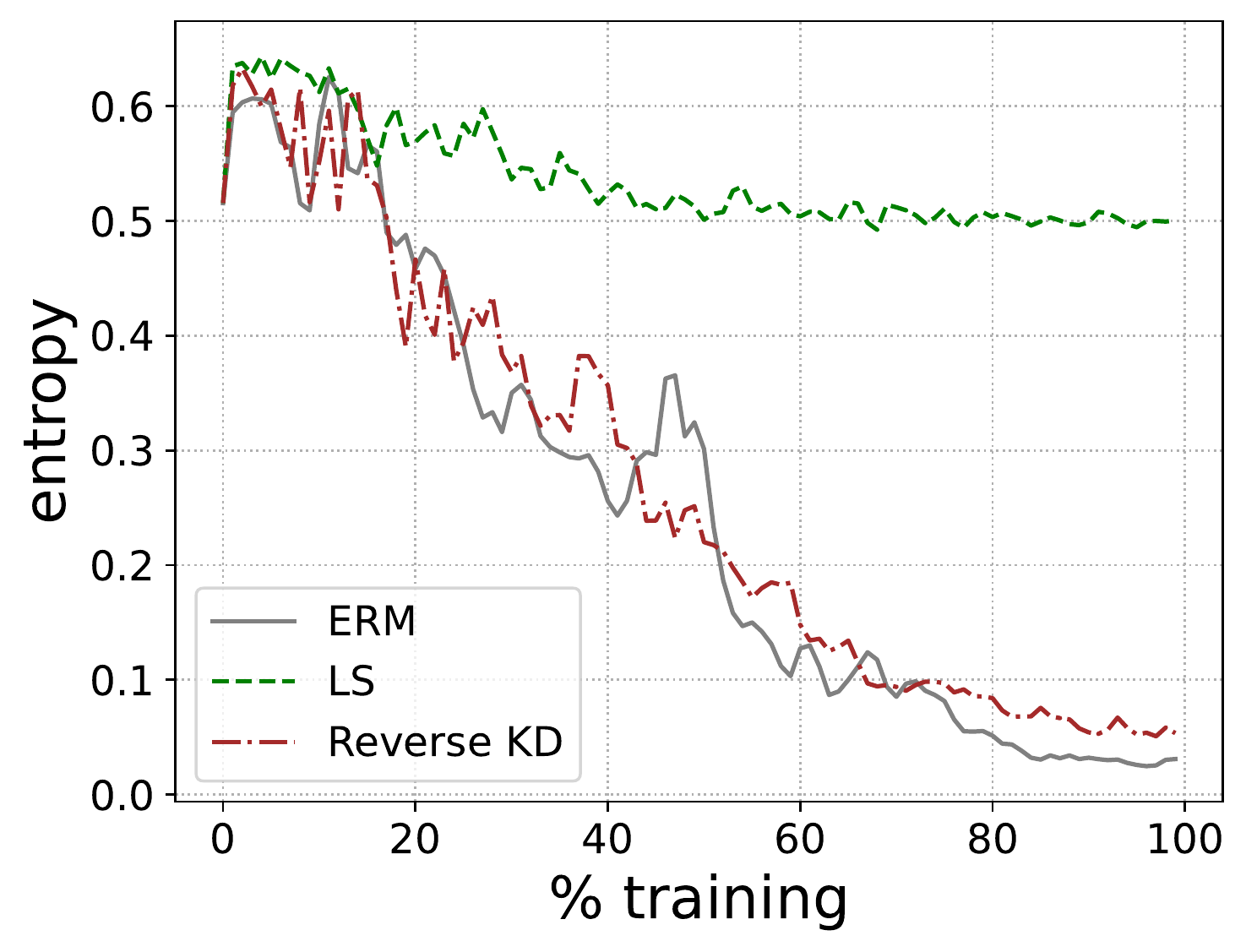}}
\subfigure[\qnli{}]{\label{figure:qnli-bstud-t1}\includegraphics[width=39mm]{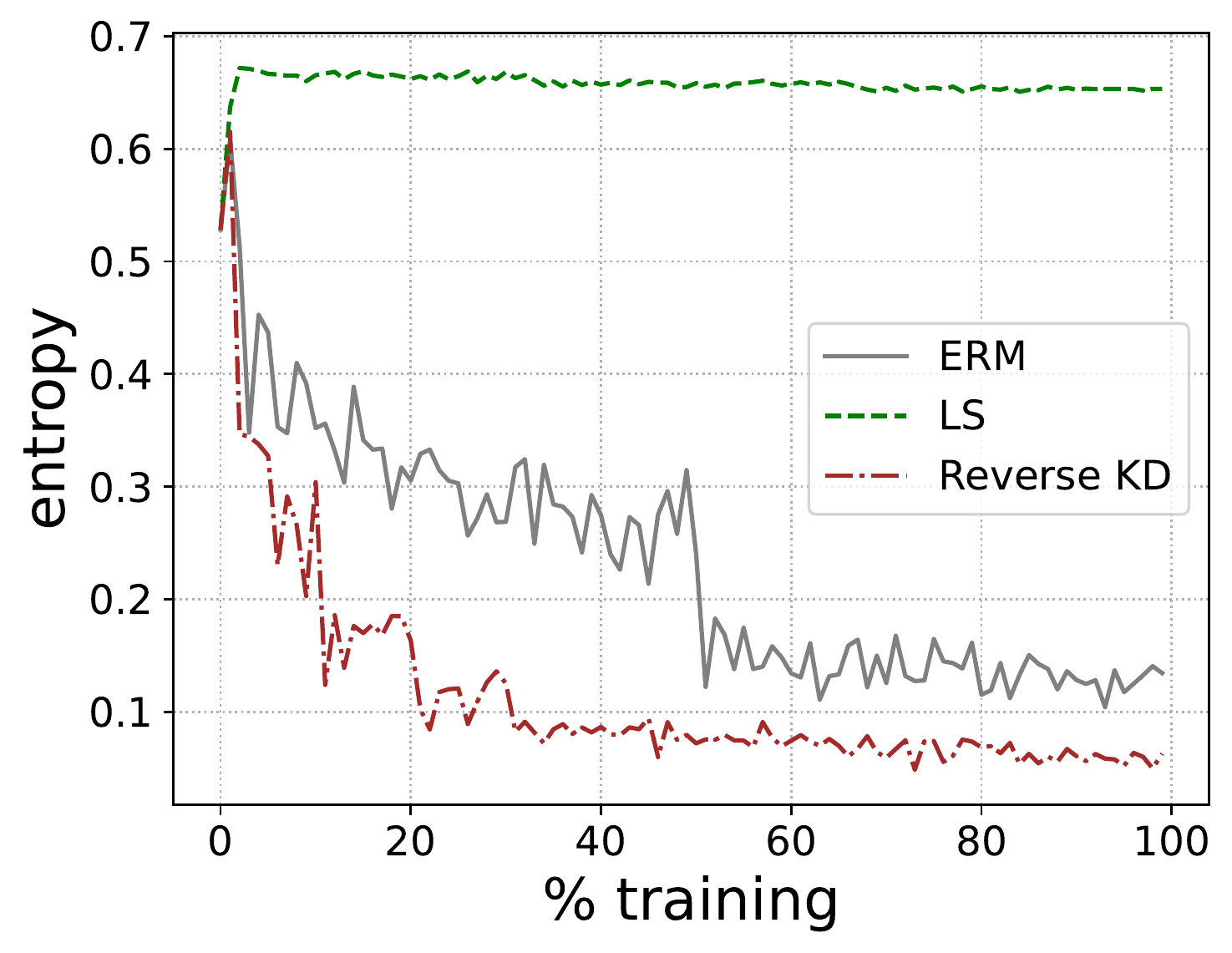}}
\subfigure[\sst{}]{\label{figure:sst2-bstud-t1}\includegraphics[width=39mm]{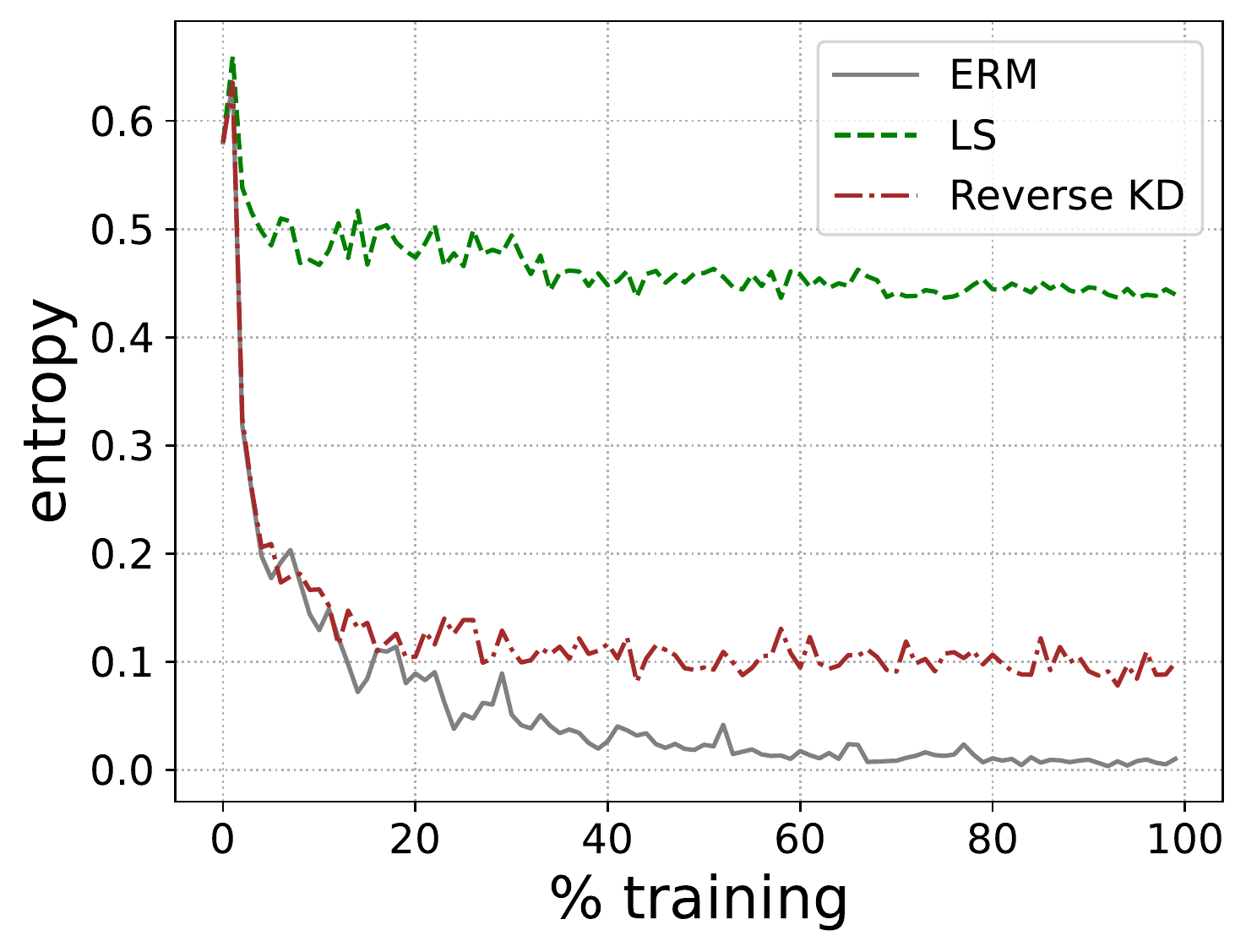}}
\subfigure[\mnli{}]{\label{figure:mnli-bstud-t1}\includegraphics[width=39mm]{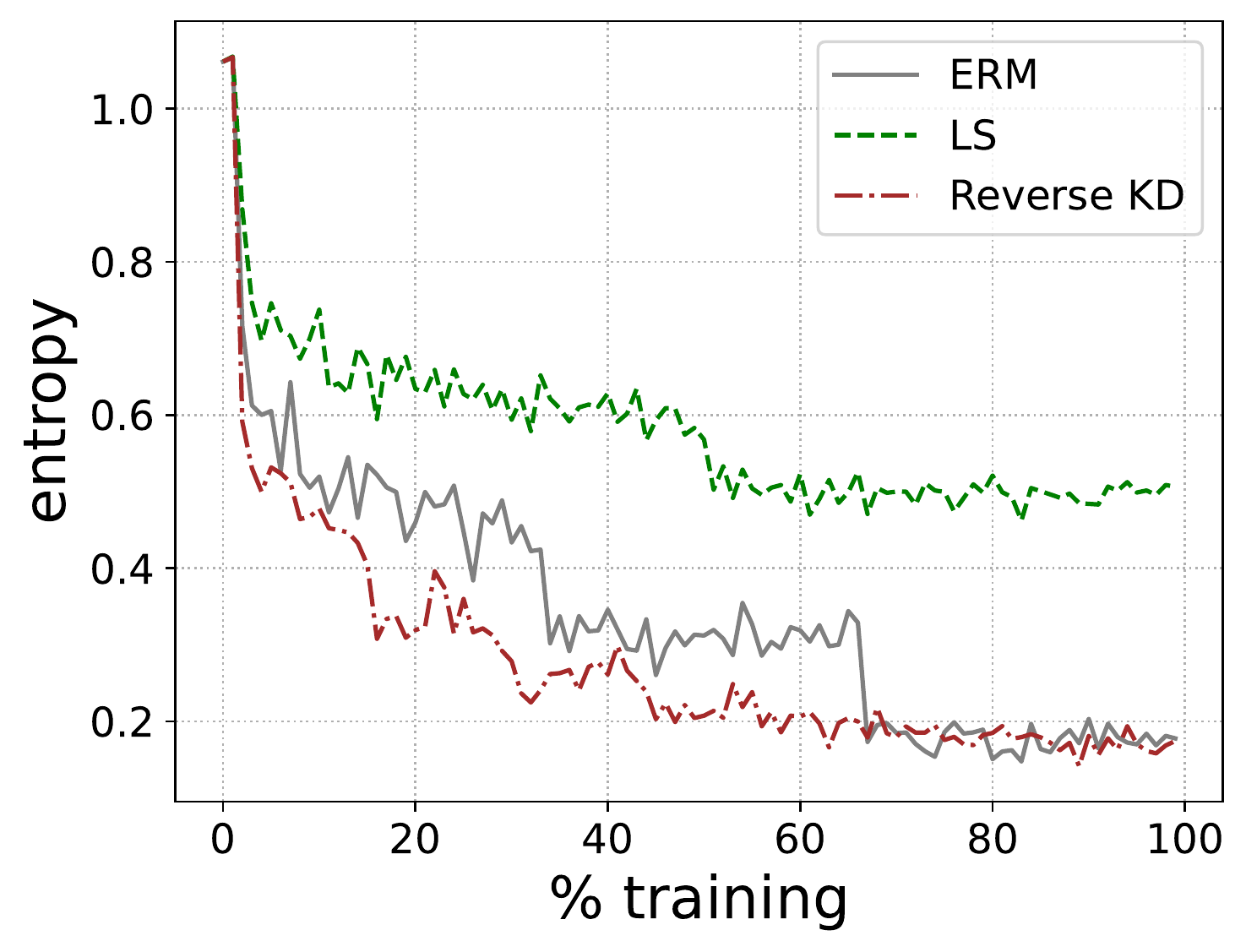}}
\caption{Change in posterior entropy of different \bert{} classifiers as training progresses.}
\label{figure:entropies-bstud-temp1}
\end{figure*}

Figure~\ref{figure:dbstud-entropies-temp1} illustrates how each training method regulates model uncertainty at real time during training.
Both \kd{} variants cause a rapid drop in entropy early in training, and an entropy lower than that of both \erm{} and \ls{} is maintained throughout.
The comparisons are also very similar between the training and the test set, indicating that the smoothing effects (or lack thereof) of each of these methods generalize well from training to inference. 

\begin{table}[t]
    \small
    \centering
    \begin{tabular}{c|ccccc}
         \textbf{Method} & \textbf{\mrpc{}} & \textbf{\qnli{}} & \textbf{\sst{}} & \textbf{\mnli{}} \\
         \hline
         \erm{} & 81.0 & 88.5 & 93.8 & 81.3 \\
         \hline
         \ls{} & \colorred{80.3} & \colorred{88.4} & \colorgreen{95.0} & \colorred{81.0} \\
         Standard \kd{} & \colorgreen{82.0} & \colorgreen{89.2} & \colorgreen{94.6} & \colorgreen{81.8} \\
         Self-\kd{} & \colorgreen{82.3} & \colorgreen{89.0} & \colorgreen{94.7} & \colorgreen{82.2} \\
    \end{tabular}
    \caption{Test set accuracy of \distilbert{} fine-tuned with \erm{}, \kd{} and \ls{}. \kd{} has the best performance profile.}
    \label{table:dbstud-performance-temp1}
\end{table}

In Table~\ref{table:dbstud-performance-temp1} we report the test set accuracy of each method on the four tasks.
Both \kd{} variants do generally better than \ls{}, as the latter also lags behind the \erm{} baseline in three out of four evaluations.
Interestingly, self-\kd{} outperforms standard \kd{} in most cases.
While \citet{yuan2020revisiting} explain similar successes of weaker teacher models in image classification with the equivalence of \kd{} and \ls{}, an alternative explanation is that these results are simply an outcome of the standard-\kd{} student not adequately learning to mimic its higher-capacity teacher due to limited amounts of training data.
Others \cite{liang2021mixkd,sultan2022overfit} have shown that such teacher-student knowledge gaps can be reduced by using additional synthetic data during distillation.

\begin{table}[t]
    \small
    \centering
    \begin{tabular}{c|ccccc}
         \textbf{Method} & \textbf{\mrpc{}} & \textbf{\qnli{}} & \textbf{\sst{}} & \textbf{\mnli{}} \\
         \hline
         \erm{} & .0678 & .1218 & .0124 & .1845 \\
         \hline
         \ls{} & \colorgreen{.5162} & \colorgreen{.6528} & \colorgreen{.4376} & \colorgreen{.4770} \\
         Reverse \kd{} & \colorgreen{.0954} & \colorred{.0834} & \colorgreen{.0954} & \colorgreen{.2041} \\
    \end{tabular}
    \caption{Test set posterior entropies of \bert{} classifiers. \kd{} increases entropy in three out of four tasks, but the \ls{}-induced increases are far greater in magnitude.
    }
    \label{table:bstud-entropies-temp1}
\end{table}

Next we turn our attention to reverse \kd{} from \distilbert{} into \bert{}.
In Table~\ref{table:bstud-entropies-temp1}, \ls{} unconditionally increases posterior entropy as before.
The effect of reverse \kd{}, on the other hand, is qualitatively different than standard and self-\kd{}, as it also increases entropy in most cases over \erm{}.
We also observe a similar difference in runtime training entropy, as depicted in Figure~\ref{figure:entropies-bstud-temp1}.

The contrast in the results of Tables \ref{table:dbstud-entropies-temp1} and \ref{table:bstud-entropies-temp1}
paints a clear and important picture: the predictive uncertainty of the student in \kd{} is a direct function of that of the teacher, and not of the \kd{} process.
To elaborate, we first point the reader to row 1 of both tables, which show that when trained using \erm{}, \bert{} classifiers have a lower posterior entropy than their smaller \distilbert{} counterparts.
Consequently, in Table~\ref{table:dbstud-entropies-temp1}, standard \kd{} yields lower-entropy students than self-\kd{}.
For the same reason, reverse \kd{} 
in Table~\ref{table:bstud-entropies-temp1} also generally trains higher-entropy models than \erm{}.

The above results have key implications for the general question concerning the true nature of \kd{}, namely, is it knowledge transfer or regularization?
As a student learns to mimic its teacher in \kd{}, we see in the above results that it also inherits its confidence---an integral aspect of the data-driven knowledge of both models.
This is clearly unlike methods such as \ls{} or teacher-free \kd{} in \cite{yuan2020revisiting}, which unconditionally reduce confidence in a data-agnostic way to limit overfitting.

With a weaker teacher model, reverse \kd{} yields a smaller overall gain in accuracy than before over \erm{} and \ls{} (see Table~\ref{table:bstud-performance-temp1}).
It is interesting, nevertheless, that a lower inductive bias along with strong language modeling capabilities enables the reverse-\kd{} student to outperform its teacher.

\begin{table}[t]
    \small
    \centering
    \begin{tabular}{c|ccccc}
         \textbf{Method} & \textbf{\mrpc{}} & \textbf{\qnli{}} & \textbf{\sst{}} & \textbf{\mnli{}} \\
         \hline
         \erm{} & 84.0 & 91.0 & 94.5 & 83.7 \\
         \hline
         \ls{} & \colorgreen{84.1} & \colorred{89.8} & \colorgreen{94.6} & 83.7 \\
         Reverse \kd{} & \colorred{83.4} & \colorgreen{91.3} & \colorgreen{94.9} & \colorgreen{83.9} \\
    \end{tabular}
    \caption{Test set accuracy of \bert{} classifiers fine-tuned with different training methods. \kd{} and \ls{} again have generally dissimilar performance profiles.}
    \label{table:bstud-performance-temp1}
\end{table}

\section{Conclusion}
\label{section:conclusion}
Knowledge distillation is a powerful framework for training lightweight and fast yet high-accuracy models.
While still a topic of active interest, any misinterpretation, including false equivalences with other methods, can hinder progress for the study of this promising approach going forward.
In this paper, we present evidence against its interpretation as label smoothing, and more generally, regularization.
We hope that our work will inspire future efforts to understand distillation better and further improve its utility.

\section*{Limitations}
To manage the complexity of our already large-scale experiments involving (a) four different classification tasks, and (b) hyperparameter grids containing up to 1,960 search points, we ran experiments with a single random seed.
While this is sufficient for exploring our main research question involving the posterior entropy of different models---for which we only need to show one counterexample, \ie{}, one where entropy decreases with knowledge distillation---the exact accuracy figures would be more reliable if averaged over multiple seeds.


\bibliography{custom}
\bibliographystyle{acl_natbib}

\clearpage
\appendix

\section{Appendix}
\label{section:appendix}
\subsection{Datasets Used in Our Experiments}
\label{subsection:appendix-datasets}
We collect all datasets and their splits from the Hugging Face Datasets library\footnote{\url{https://huggingface.co/docs/datasets/index}}.
Among the four tasks, only \mrpc{} has a labeled test set.
For the other three datasets, we split the train set into two parts.
In \qnli{} and \mnli{} experiments, the two parts are used for training and validation, and the original validation set is used as a blind test set.
For \sst{}, since the validation set is small, we also use it for model selection; the train set is therefore split into train and test sets.
For \mnli{}, we evaluate only on the \texttt{test-matched} set, which consists of unseen in-domain examples.
Table~\ref{table:datasets} shows the number of examples in each dataset.

\begin{table}[h]
    \small
    \centering
    \begin{tabular}{c|ccc}
         \textbf{Task} & \textbf{Train} & \textbf{Validation} & \textbf{Test} \\
         \hline
         \mrpc{} & 3,668 & 408 & 1,725 \\
         \qnli{} & 99,743 & 5,000 & 5,463 \\
         \sst{} & 65,549 & 872 & 1,800 \\
         \mnli{} & 382,887 & 9,815 & 9,815 \\
         
    \end{tabular}
    \caption{Size statistics for our datasets.}
    \label{table:datasets}
\end{table}

\subsection{Model Selection}
\label{subsection:appendix-optimal-hp}
Table~\ref{table:optimal-hp-tau1} contains the optimal configurations for the models evaluated in \S{\ref{section:experiments}} as observed on the respective validations sets.

\subsection{Distillation at Higher Temperatures}
\label{subsection:appendix-high-temp-kd}
We present results for the same experiments as in \S{\ref{section:experiments}} here, with two main differences:
First, for \kd{}, we include the temperature $\tau$ in the hyperparameter search and re-define the domain for $\alpha$ as follows:
\begin{table}[h!]
    \small
    \centering
    \begin{tabular}{cc}
        \textbf{Hyperparameter} & \textbf{Values} \\
        \hline
        $\tau$ & $\{1, 2, 4, 8, 16, 32, 64\}$ \\
        $\alpha$ (Eq. \ref{equation:kd-cost-function}) & $\{.25, .5, .75, 1\}$ \\
        Learning rate & $\{1, 2, ..., 7\} \times 10^{-5}$ \\
        $\#$ of epochs & $\{1, 2, ..., 10\}$
    \end{tabular}
    \label{table:hp-grid-ht}
\end{table}
~\\[-8pt]
\noindent
yielding a 4D grid consisting of 1,960 points. Note that for this more general experiment, we allow $\alpha=1$, which only uses the teacher's soft targets for training.
Second, to keep the grid size the same (1,960) for \ls{}, we re-define the domain of $\alpha$ for \ls{} to contain 28 equally-distanced points ranging from .1 to .9991.

\begin{figure*}[t]
\centering
\subfigure[\mrpc{}]{\label{figure:mrpc-dbstud}\includegraphics[width=39mm]{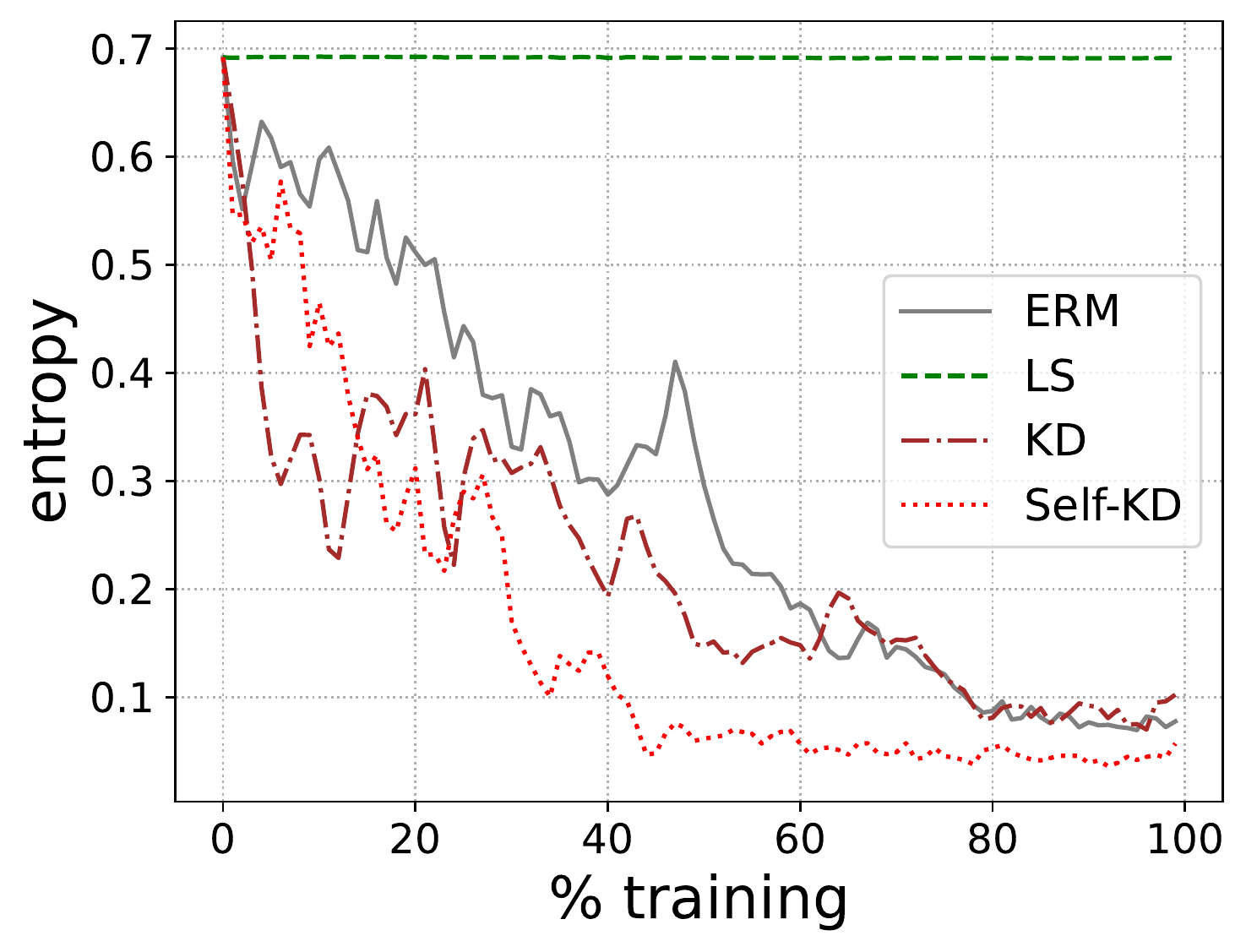}}
\subfigure[\qnli{}]{\label{figure:qnli-dbstud}\includegraphics[width=39mm]{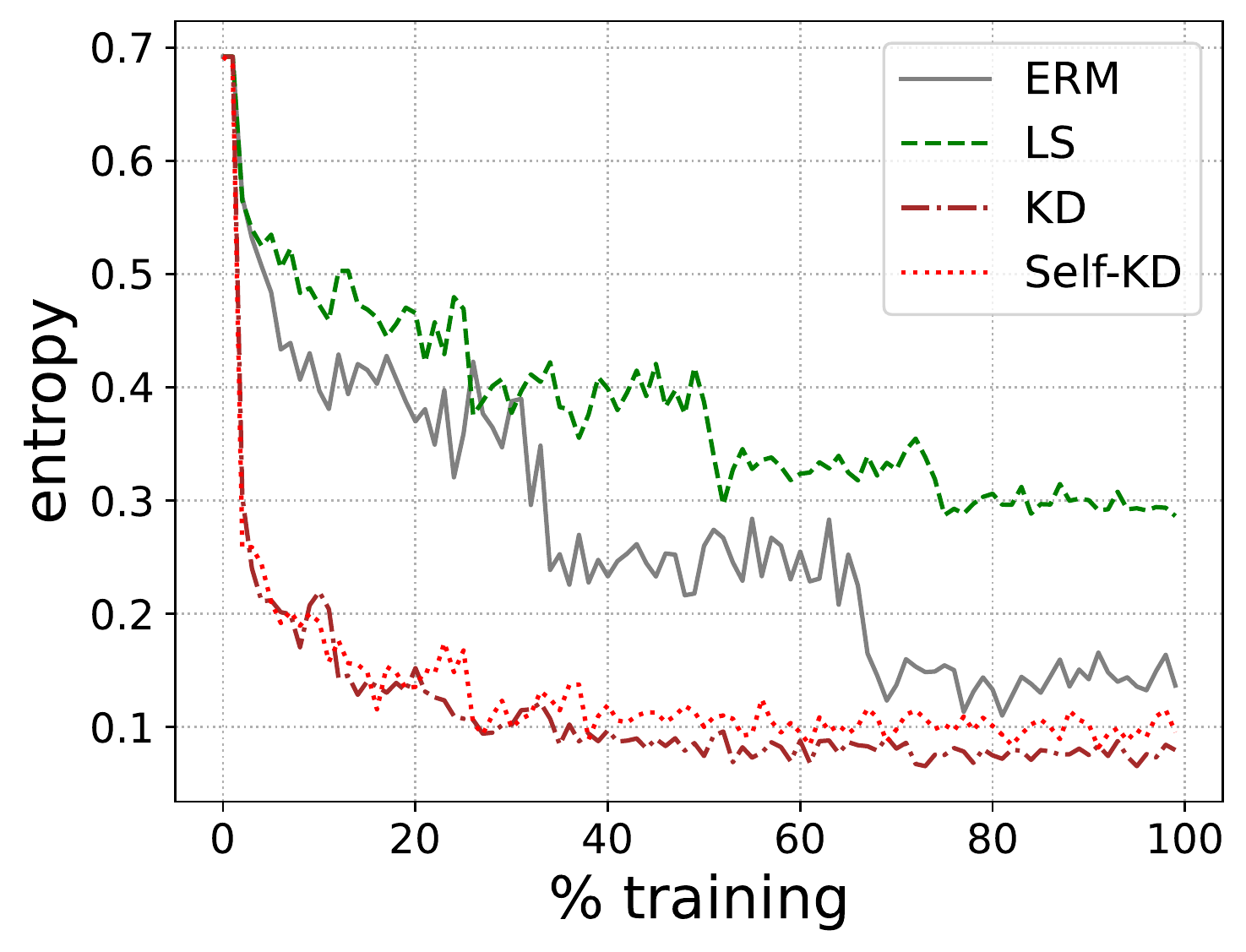}}
\subfigure[\sst{}]{\label{figure:sst2-dbstud}\includegraphics[width=39mm]{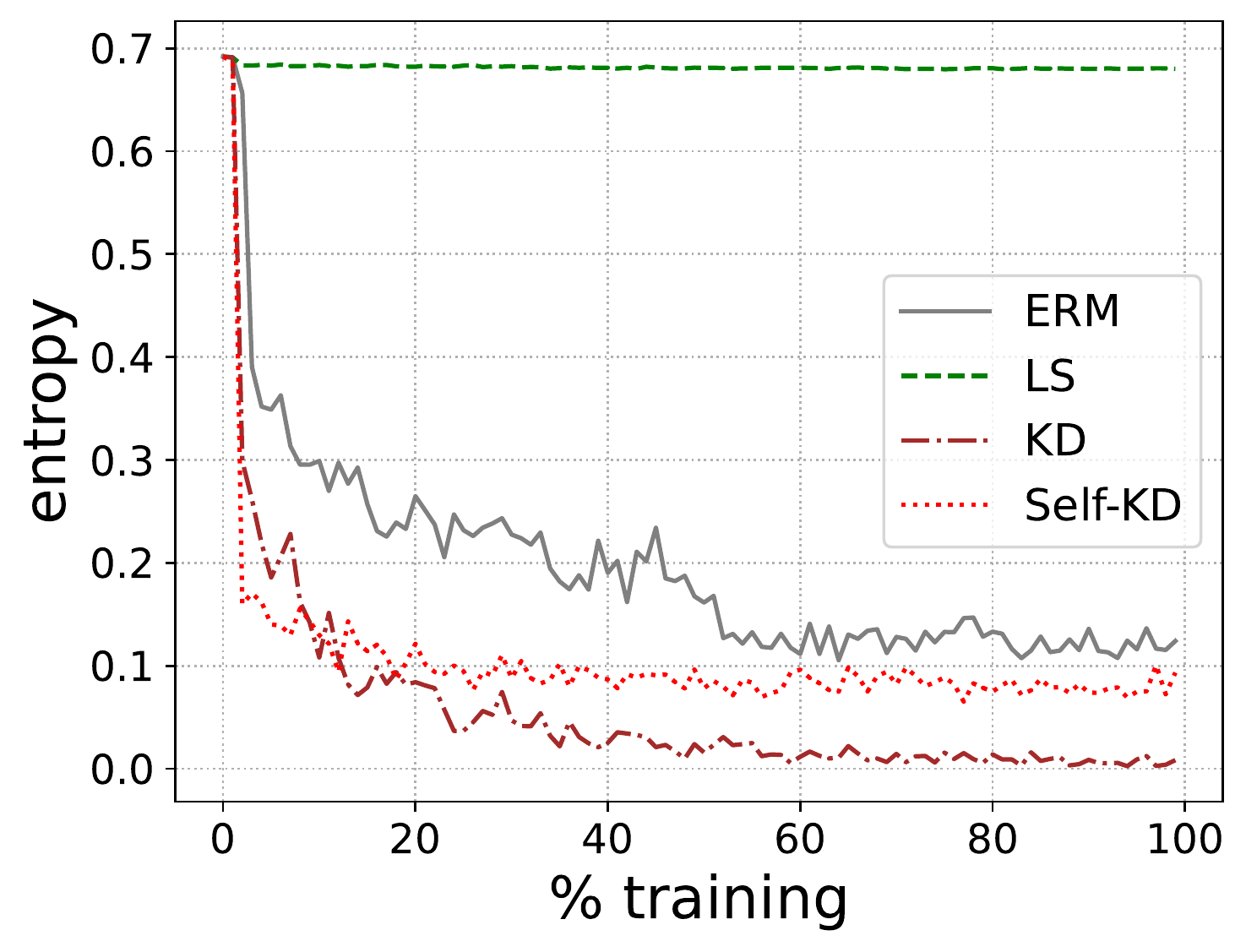}}
\subfigure[\mnli{}]{\label{figure:mnli-dbstud}\includegraphics[width=39mm]{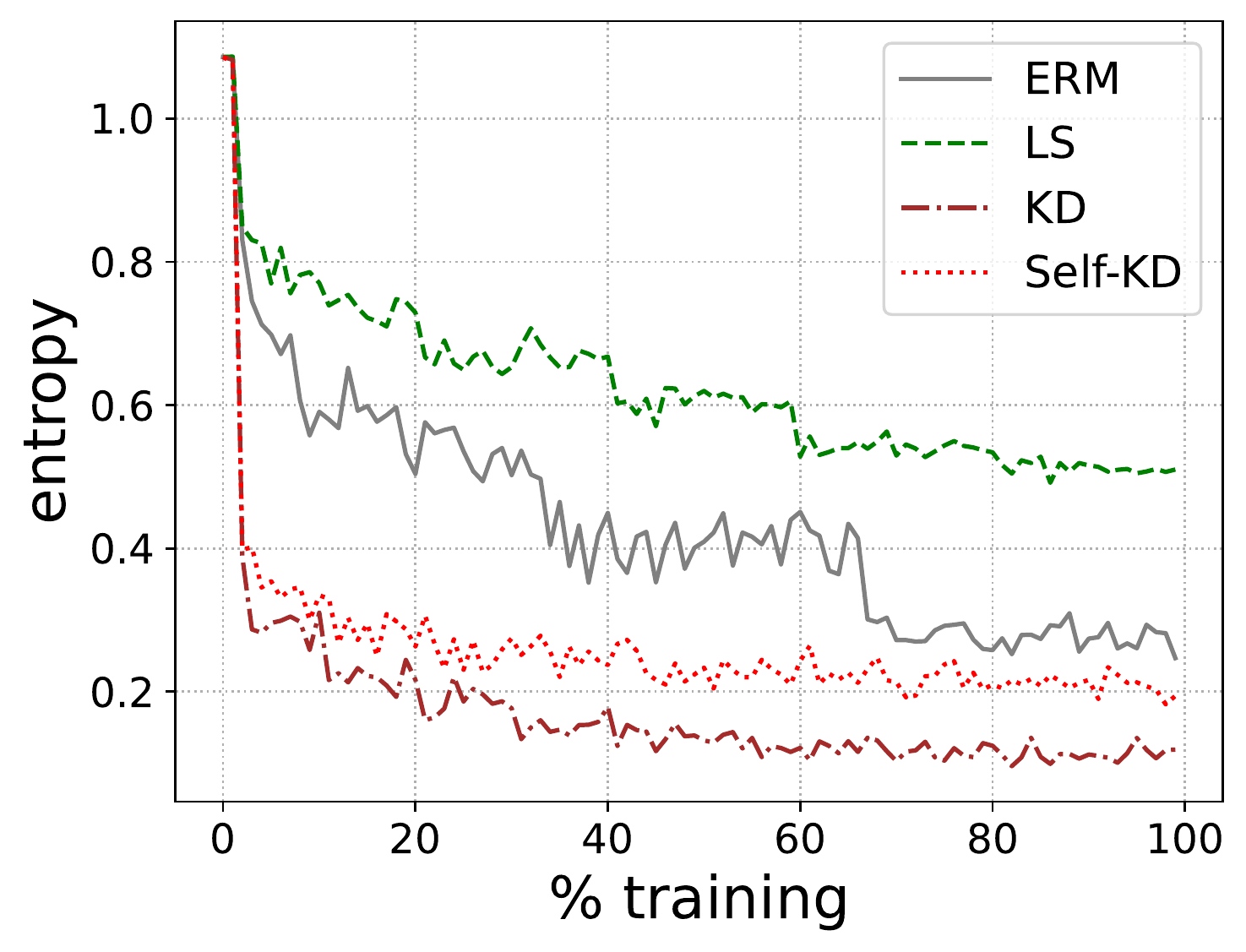}}
\caption{Change in posterior entropy for different \distilbert{} classifiers as training (at optimal $\tau$) progresses.}
\label{figure:dbstud-entropies}
\end{figure*}

In Table~\ref{table:dbstud-entropies} and Figure~\ref{figure:dbstud-entropies}, we report posterior entropies of \distilbert{} models measured at inference and during training, respectively.
Similar to the experimental results in \S{\ref{section:experiments}}, both \kd{} variants generally train models with lower entropy than not only \ls{}, but also \erm{}.
In Table~\ref{table:dbstud-performance}, \kd{} again has better overall test set accuracy than \ls{}.
Across all these results, the comparisons between standard and self-\kd{} also remain generally unchanged.

\begin{figure*}[t]
\centering
\subfigure[\mrpc{}]{\label{figure:mrpc-bstud}\includegraphics[width=39mm]{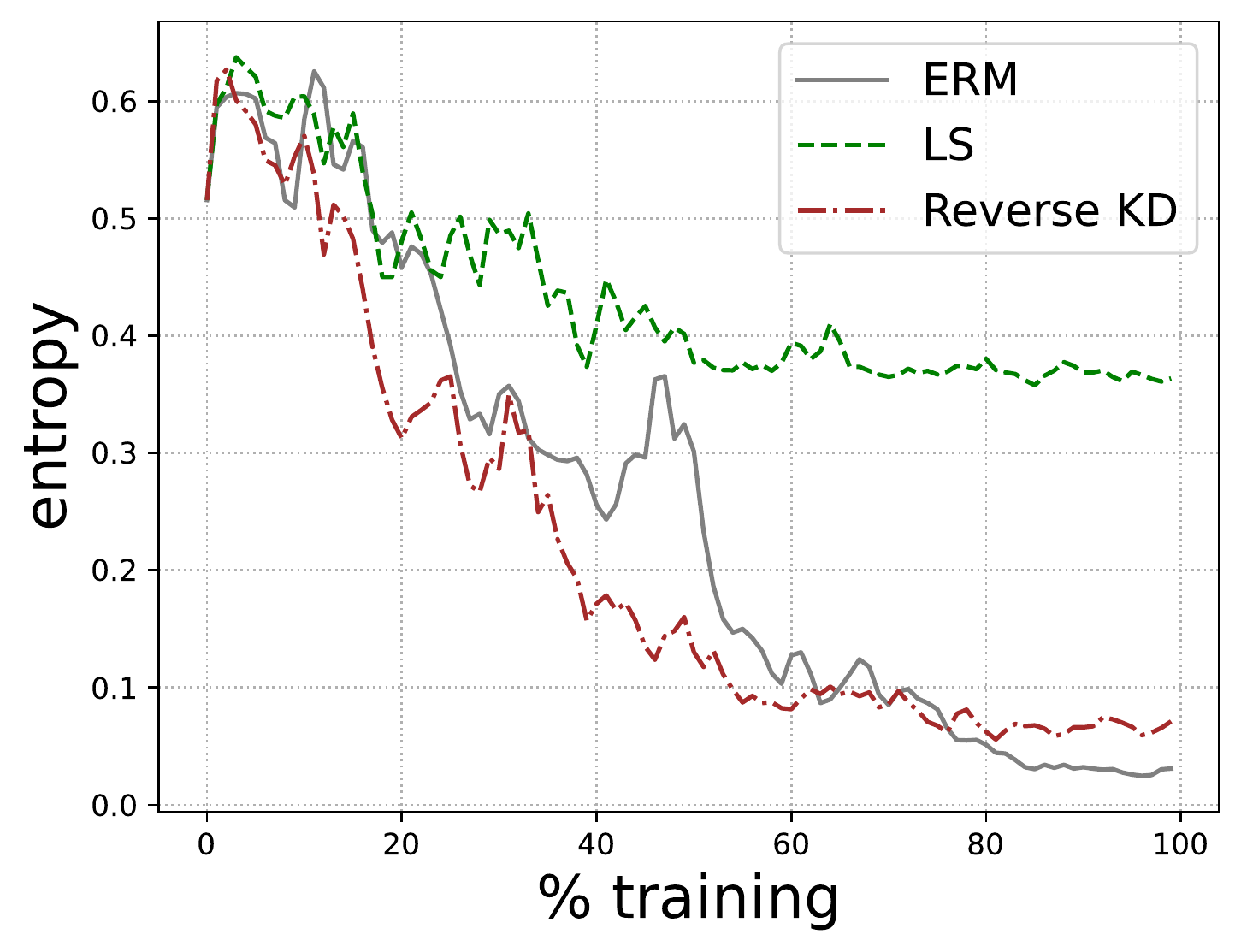}}
\subfigure[\qnli{}]{\label{figure:qnli-bstud}\includegraphics[width=39mm]{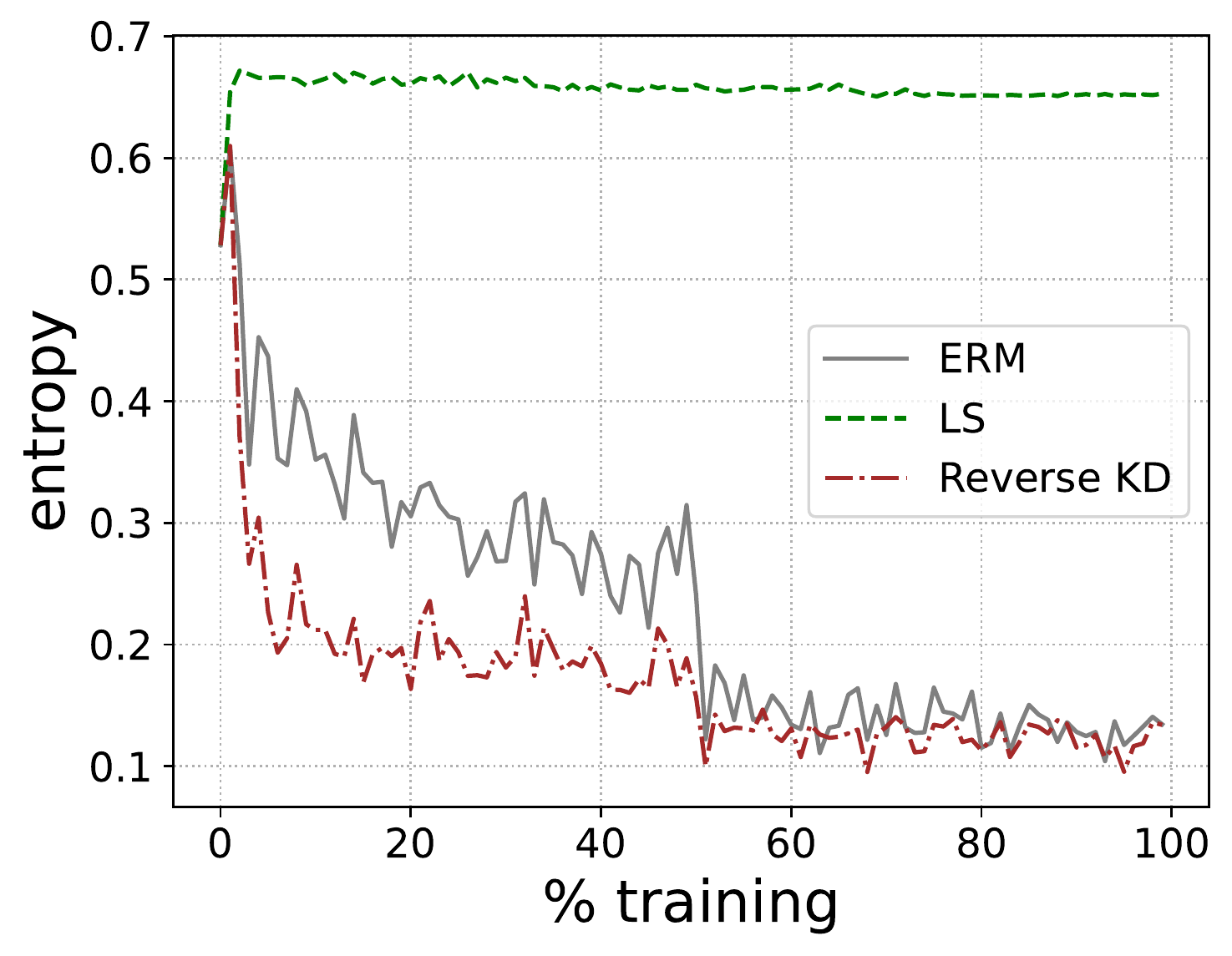}}
\subfigure[\sst{}]{\label{figure:sst2-bstud}\includegraphics[width=39mm]{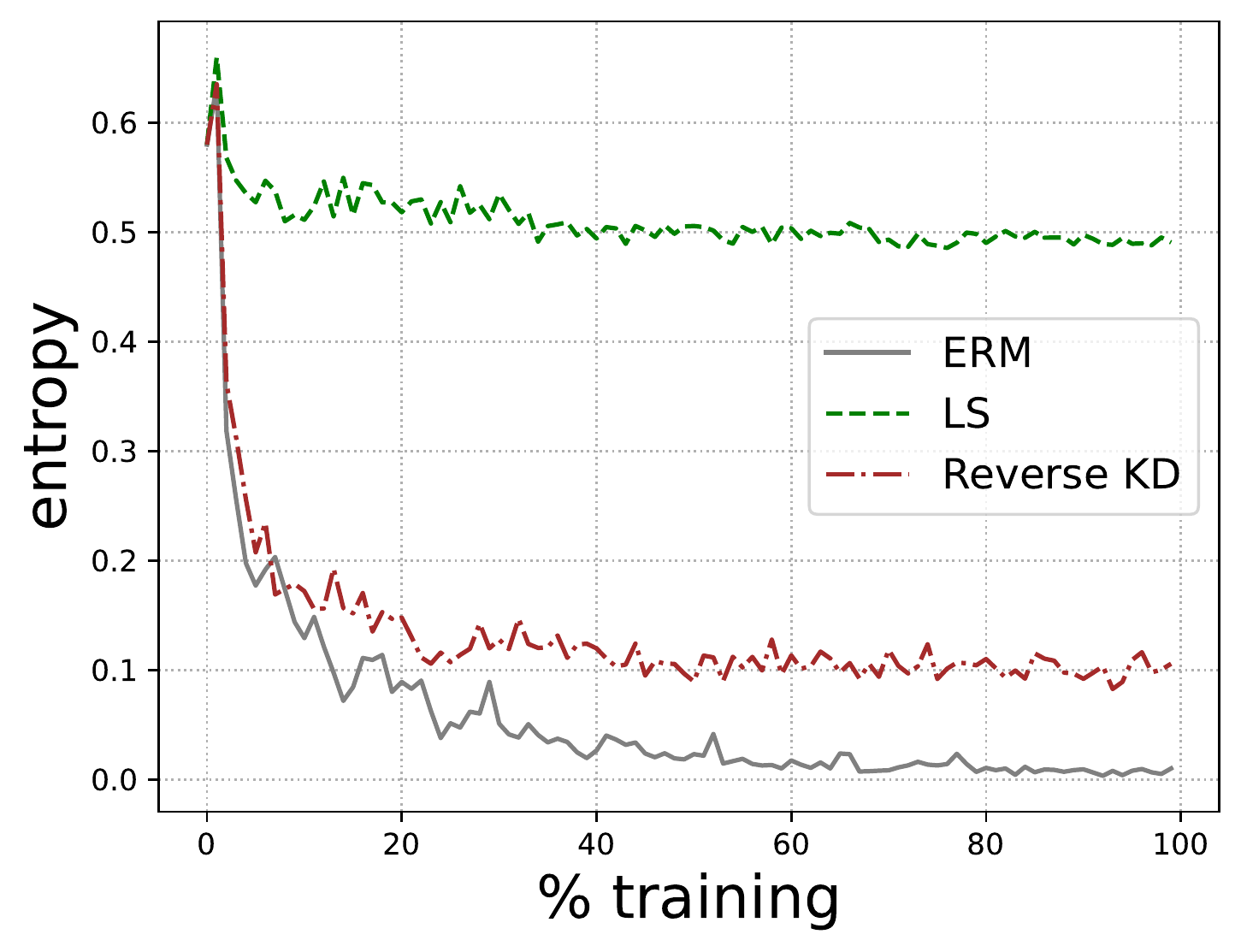}}
\subfigure[\mnli{}]{\label{figure:mnli-bstud}\includegraphics[width=39mm]{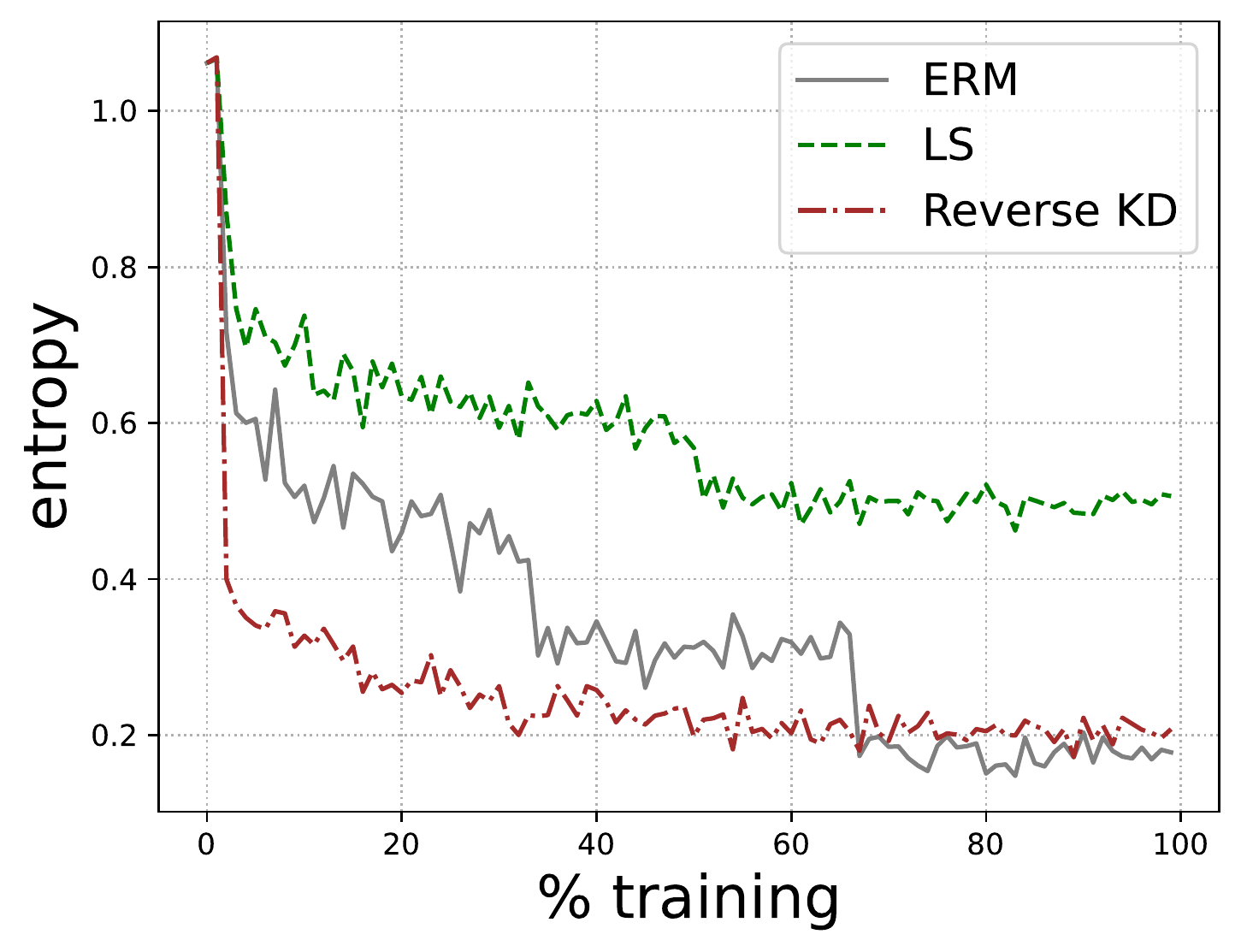}}
\caption{Change in posterior entropy for different \bert{} classifiers as training (at optimal $\tau$) progresses.}
\label{figure:bstud-entropies}
\end{figure*}

The reverse \kd{} results---for both predictive uncertainty in Table~\ref{table:bstud-entropies} and Figure~\ref{figure:bstud-entropies}, and test set performance in Table~\ref{table:bstud-performance}---also show very similar trends as those described in \S{\ref{section:experiments}}.
Altogether, the results presented in this section with \kd{} in three different settings demonstrate that even at high temperatures, our original findings and conclusions hold true.

The optimal hyperparameter combinations for the above experiments are provided in Table~\ref{table:optimal-hp}.

\begin{table*}[t]
    \small
    \centering
    \begin{tabular}{c|c|c|ccc}
        \hline
        \textbf{Task} & \textbf{Model} & \textbf{Training} & $\boldsymbol{\alpha}$ & \textbf{Learning Rate} & \textbf{$\#$ Epochs} \\
        \hline
        \multirow{7}{*}{\mrpc{}} & \multirow{3}{*}{\bert{}} & \erm{}         & $-$ & 7e-5 & 4 \\
                &                                           & \ls{}          & .4  & 7e-5 & 7 \\
                &                                           & Reverse \kd{}  & .1  & 4e-5 & 5 \\
            \cline{2-6}
                & \multirow{4}{*}{\distilbert{}}            & \erm{}         & $-$ & 5e-5 & 4 \\
                &                                           & \ls{}          & .9  & 4e-5 & 4 \\
                &                                           & Standard \kd{} & .1  & 6e-5 & 7 \\
                &                                           & Self-\kd{}     & .8  & 7e-5 & 6 \\
        \hline
        \multirow{7}{*}{\qnli{}} & \multirow{3}{*}{\bert{}} & \erm{}         & $-$ & 4e-5 & 2 \\
                &                                           & \ls{}          & .7  & 2e-5 & 3 \\
                &                                           & Reverse \kd{}  & .3  & 2e-5 & 10 \\
            \cline{2-6}
                & \multirow{4}{*}{\distilbert{}}            & \erm{}         & $-$ & 3e-5 & 3 \\
                &                                           & \ls{}          & .8  & 2e-5 & 8 \\
                &                                           & Standard \kd{} & .8  & 3e-5 & 5 \\
                &                                           & Self-\kd{}     & .7  & 2e-5 & 10 \\
        \hline
        \multirow{7}{*}{\sst{}} & \multirow{3}{*}{\bert{}} & \erm{}         & $-$  & 1e-5 & 9 \\
                &                                           & \ls{}          & .3  & 2e-5 & 3 \\
                &                                           & Reverse \kd{}  & .7  & 1e-5 & 7 \\
            \cline{2-6}
                & \multirow{4}{*}{\distilbert{}}            & \erm{}         & $-$ & 1e-5 & 2 \\
                &                                           & \ls{}          & .3  & 6e-5 & 2 \\
                &                                           & Standard \kd{} & .6  & 6e-5 & 2 \\
                &                                           & Self-\kd{}     & .7  & 5e-5 & 8 \\
        \hline
        \multirow{7}{*}{\mnli{}} & \multirow{3}{*}{\bert{}} & \erm{}         & $-$  & 3e-5 & 3 \\
                &                                           & \ls{}          & .1  & 3e-5 & 2 \\
                &                                           & Reverse \kd{}  & .4  & 3e-5 & 7 \\
            \cline{2-6}
                & \multirow{4}{*}{\distilbert{}}            & \erm{}         & $-$ & 3e-5 & 3 \\
                &                                           & \ls{}          & .8  & 3e-5 & 3 \\
                &                                           & Standard \kd{} & .7  & 3e-5 & 6 \\
                &                                           & Self-\kd{}     & .9  & 3e-5 & 9 \\
        \hline
    \end{tabular}
    \caption{Optimal hyperparameter combinations ($\tau$=1) for the \S{\ref{section:experiments}} experiments.}
    \label{table:optimal-hp-tau1}
\end{table*}

\begin{table}[h]
    \small
    \centering
    \begin{tabular}{c|ccccc}
         \textbf{Method} & \textbf{\mrpc{}} & \textbf{\qnli{}} & \textbf{\sst{}} & \textbf{\mnli{}} \\
         \hline
         \erm{} & .1123 & .1419 & .1101 & .2698 \\
         \hline
         \ls{} & \colorgreen{.6916} & \colorgreen{.2949} & \colorgreen{.6802} & \colorgreen{.5095} \\
         Standard \kd{} & \colorred{.1067} & \colorred{.1024} & \colorred{.0107} & \colorred{.1481} \\
         Self-\kd{} & \colorred{.0927} & \colorred{.1179} & \colorred{.0853} & \colorred{.2476} \\         
    \end{tabular}
    \caption{Test set posterior entropies of \distilbert{} classifiers fine-tuned with different training methods. Label smoothing (\ls{}) increases entropy over \erm{} as expected, but distillation (\kd{}) induces a drop on all tasks.}
    \label{table:dbstud-entropies}
\end{table}

\begin{table}[h]
    \small
    \centering
    \begin{tabular}{c|ccccc}
         \textbf{Method} & \textbf{\mrpc{}} & \textbf{\qnli{}} & \textbf{\sst{}} & \textbf{\mnli{}} \\
         \hline
         \erm{} & 81.0 & 88.5 & 93.8 & 81.3 \\
         \hline
         \ls{} & \colorred{78.4} & \colorgreen{88.8} & \colorgreen{95.2} & \colorred{80.8} \\
         Standard \kd{} & \colorred{80.2} & \colorgreen{89.6} & \colorgreen{94.6} & \colorgreen{82.7} \\
         Self-\kd{} & \colorgreen{82.1} & \colorgreen{89.5} & \colorgreen{94.8} & \colorgreen{82.6} \\
    \end{tabular}
    \caption{Test set accuracy of \distilbert{} classifiers fine-tuned with different training methods. \kd{} has the best overall performance profile.}
    \label{table:dbstud-performance}
\end{table}

\begin{table}[h]
    \small
    \centering
    \begin{tabular}{c|ccccc}
         \textbf{Method} & \textbf{\mrpc{}} & \textbf{\qnli{}} & \textbf{\sst{}} & \textbf{\mnli{}} \\
         \hline
         \erm{} & .0678 & .1218 & .0124 & .1844 \\
         \hline
         \ls{} & \colorgreen{.3778} & \colorgreen{.6512} & \colorgreen{.4891} & \colorgreen{.4770} \\
         Reverse \kd{} & \colorgreen{.1058} & \colorred{.1140} & \colorgreen{.0995} & \colorgreen{.2163} \\
    \end{tabular}
    \caption{Test set posterior entropies of \bert{} classifiers fine-tuned with different training methods.
    While \kd{} does increase entropy over \erm{} on three out of four tasks, \ls{} yields increases that are much higher.}
    \label{table:bstud-entropies}
\end{table}

\begin{table}[h]
    \small
    \centering
    \begin{tabular}{c|ccccc}
         \textbf{Method} & \textbf{\mrpc{}} & \textbf{\qnli{}} & \textbf{\sst{}} & \textbf{\mnli{}} \\
         \hline
         \erm{} & 84.0 & 91.0 & 94.5 & 83.7\\
         \hline
         \ls{} & \colorred{83.9} & \colorred{90.8} & 94.5 & 83.7 \\
         Reverse \kd{} & \colorgreen{84.3} & \colorgreen{91.3} & \colorgreen{94.9} & \colorgreen{84.0}
    \end{tabular}
    \caption{Test set accuracy of \bert{} classifiers fine-tuned with different training methods. \kd{} clearly outperforms \ls{}.}
    \label{table:bstud-performance}
\end{table}

\begin{table*}[t]
    \small
    \centering
    \begin{tabular}{c|c|c|cccc}
        \hline
        \textbf{Task} & \textbf{Model} & \textbf{Training} & $\boldsymbol{\alpha}$ & $\boldsymbol{\tau}$ & \textbf{Learning Rate} & \textbf{$\#$ Epochs} \\
        \hline
        \multirow{7}{*}{\mrpc{}} & \multirow{3}{*}{\bert{}} & \erm{}         & $-$    & $-$ & 7e-5 & 4 \\
                &                                           & \ls{}          & .2332  & $-$ & 7e-5 & 6 \\
                &                                           & Reverse \kd{}  & 1      & 1    & 4e-5  & 6 \\
            \cline{2-7}
                & \multirow{4}{*}{\distilbert{}}            & \erm{}         & $-$    & $-$  & 5e-5 & 4 \\
                &                                           & \ls{}          & .9325  & $-$  & 3e-5 & 5 \\
                &                                           & Standard \kd{} & 1      & 16   & 3e-5 & 3 \\
                &                                           & Self-\kd{}     & .25    & 2    & 7e-5 & 7 \\
        \hline
        \multirow{7}{*}{\qnli{}} & \multirow{3}{*}{\bert{}} & \erm{}         & $-$    & $-$  & 4e-5 & 2 \\
                &                                           & \ls{}          & .6994  & $-$  & 4e-5 & 3 \\
                &                                           & Reverse \kd{}  & .25    & 64   & 4e-5 & 2 \\
            \cline{2-7}
                & \multirow{4}{*}{\distilbert{}}            & \erm{}         & $-$    & $-$  & 3e-5 & 3 \\
                &                                           & \ls{}          & .1333  & $-$  & 3e-5 & 4 \\
                &                                           & Standard \kd{} & .75    & 64   & 3e-5 & 9 \\
                &                                           & Self-\kd{}     & .5     & 64   & 3e-5 & 8 \\
        \hline
        \multirow{7}{*}{\sst{}} & \multirow{3}{*}{\bert{}} & \erm{}          & $-$    & $-$  & 1e-5 & 9 \\
                &                                           & \ls{}          & .3664  & $-$  & 2e-5 & 3 \\
                &                                           & Reverse \kd{}  & .75    & 1    & 1e-5 & 5 \\
            \cline{2-7}
                & \multirow{4}{*}{\distilbert{}}            & \erm{}         & $-$    & $-$  & 1e-5 & 2 \\
                &                                           & \ls{}          & .8326  & $-$  & 5e-5 & 3 \\
                &                                           & Standard \kd{} & .5     & 1    & 2e-5 & 9 \\
                &                                           & Self-\kd{}     & .25    & 8    & 5e-5 & 6 \\
        \hline
        \multirow{7}{*}{\mnli{}} & \multirow{3}{*}{\bert{}} & \erm{}         & $-$    & $-$  & 3e-5 & 3 \\
                &                                           & \ls{}          & .1     & $-$  & 3e-5 & 2 \\
                &                                           & Reverse \kd{}  & .5     & 4    & 2e-5 & 7 \\
            \cline{2-7}
                & \multirow{4}{*}{\distilbert{}}            & \erm{}         & $-$    & $-$  & 3e-5 & 3 \\
                &                                           & \ls{}          & .1666  & $-$  & 3e-5 & 5 \\
                &                                           & Standard \kd{} & 1      & 4    & 4e-5 & 10 \\
                &                                           & Self-\kd{}     & .75    & 64   & 5e-5 & 9 \\
        \hline
    \end{tabular}
    \caption{Optimal hyperparameter combinations for the \S{\ref{subsection:appendix-high-temp-kd}} experiments.}
    \label{table:optimal-hp}
\end{table*}

\end{document}